\documentclass[10pt,twocolumn,letterpaper]{article}

\usepackage{cvpr}              % To produce the CAMERA-READY version
\usepackage{graphicx}
\usepackage{amsmath}
\usepackage{amssymb}
\usepackage{booktabs}
\usepackage[accsupp]{axessibility}  % Improves PDF readability for those with disabilities.

\usepackage[pagebackref,breaklinks,colorlinks]{hyperref}

\usepackage[capitalize]{cleveref}
\crefname{section}{Sec.}{Secs.}
\Crefname{section}{Section}{Sections}
\Crefname{table}{Table}{Tables}
\crefname{table}{Tab.}{Tabs.}

\def\cvprPaperID{3856} % *** Enter the CVPR Paper ID here
\def\confName{CVPR}
\def\confYear{2022}

\begin{document}

\title{Cross-modal Clinical Graph Transformer for Ophthalmic Report Generation}

\author{Mingjie Li$^{1,5}$\hspace{6mm}
Wenjia Cai$^2$\hspace{6mm}
Karin Verspoor$^3$\hspace{6mm}
Shirui Pan$^1$ \hspace{6mm}
Xiaodan Liang$^4$ \hspace{6mm}
Xiaojun Chang$^5$\thanks{Corresponding author.} \\
$^1$Department of Data Science and Artificial Intelligence, Monash University \\
$^2$State Key Laboratory of Ophthalmology, Zhongshan Ophthalmic Center, Sun Yat-Sen University \\
$^3$School of Computing Technologies, RMIT University \\
$^4$School of ISE, Sun Yat-Sen University, Peng Cheng National Lab \\
$^5$ReLER, AAII, University of Technology Sydney \\
}

\maketitle

%%%%%%%%% ABSTRACT
\begin{abstract}
Automatic generation of ophthalmic reports using data-driven neural networks has great potential in clinical practice. 
When writing a report, ophthalmologists make inferences with prior clinical knowledge. 
This knowledge has been neglected in prior medical report generation methods.
To endow models with the capability of incorporating expert knowledge, we propose a \textbf{C}ross-modal clinical \textbf{G}raph \textbf{T}ransformer (CGT) for ophthalmic report generation (ORG), in which clinical relation triples are injected into the visual features as prior knowledge to drive the decoding procedure.
However, two major common Knowledge Noise (KN) issues may affect models' effectiveness.
1)~Existing general biomedical knowledge bases such as the UMLS may not align meaningfully to the specific context and language of the report, limiting their utility for knowledge injection.
2)~Incorporating too much knowledge may divert the visual features from their correct meaning. 
To overcome these limitations, we design an automatic information extraction scheme based on natural language processing to obtain clinical entities and relations directly from in-domain training reports.
Given a set of ophthalmic images, our CGT first restores a sub-graph from the clinical graph and injects the restored triples into visual features.
Then visible matrix is employed during the encoding procedure to limit the impact of knowledge. 
Finally, reports are predicted by the encoded cross-modal features via a Transformer decoder. 
Extensive experiments on the large-scale FFA-IR benchmark demonstrate that the proposed CGT is able to outperform previous benchmark methods and achieve state-of-the-art performances.

\end{abstract}
\vspace{-0.5cm}
%%%%%%%%% BODY TEXT
\section{Introduction}
\label{sec:intro}

Fundus Fluorescein Angiography (FFA) is one of the essential ophthalmic imaging examinations in clinical practice. However, writing reports to summarize findings from dozens of ophthalmic images during an examination is time-consuming and error-prone, especially for inexperienced ophthalmologists. 
With the success of data-driven neural networks\cite{zhang2020gis, li2018rotated,li2019knowledge,han2020mining,Lin_2022_CVPR} in many real-life scenarios, researchers and ophthalmologists start to investigate how to apply artificial intelligent (AI) models in clinical ophthalmic practice and acquire significant achievements\cite{retinalsuvery}.
Automatic generation of ophthalmic reports offers the possibility of reducing the heavy workloads of ophthalmologists. 
Furthermore, the predicted reports can highlight abnormalities for the ophthalmologists and provide a rationale for disease diagnosis; hence, automatic ophthalmic report generation has attracted increasing research interest for AI-based clinical decision support, as well as presenting a meaningful opportunity to explore the integration of vision and language modalities in neural network models.
 
Despite significant progress in generic image captioning models\cite{chen2020say, anderson2018bottom}, when transferring them into medical knowledge-driven tasks, they fail to achieve promising and competitive performance due to a lack of prior medical knowledge. When describing ophthalmic images, ordinary people can only recognize the common visual information, such as the shape and color, while ophthalmologists make inferences with their prior clinical knowledge. 
For models to achieve this capability, recent work explores the incorporation of medical knowledge to enhance diagnostic models\cite{li2019mrg,li2020auxiliary,zhang2020radiology,liu2021exploring}.

On the one hand, researchers\cite{li2019mrg,li2020auxiliary} have explored graph structure weights as posterior knowledge to alleviate the textual bias. 
In each graph, the nodes are observed abnormalities selected from prior knowledge, such as external medical corpus, and the edges are the predicted weights correlating each pair abnormalities. However, the weight graph limits the effectiveness of the knowledge graph from two aspects. Firstly, some entities are extracted from the external medical corpus or knowledge graph database separated from the training corpus. These entities will bring in a heterogeneous embedding space\cite{liu2020k} which makes the embedding vectors inconsistent. Secondly, there are no ground truth weights to supervise the message passing procedure, and the model is still prone to be distracted by the visual bias in medical images\cite{liu2021exploring}. On the other hand, a universal graph is proposed with prior knowledge on 20 chest findings\cite{zhang2020radiology} to enhance models. Since these findings are not always depicted in one report, incorporating all this knowledge may divert the visual features from their original meaning.

To address these issues, we propose a \textbf{C}ross-modal clinical \textbf{G}raph \textbf{T}ransformer (CGT) for ophthalmic report generation (ORG). In particular, we first invoke an information extraction scheme based on a natural language processing pipeline, including named entity recognition and entity linking, to obtain a clinical knowledge graph. More details will be introduced in Section~\ref{sec:scheme}. As discussed in \cite{jain2021radgraph}, the structured clinical information behind the free-text reports can enhance the diagnostic methods. In addition, the entities and relations in our clinical graph are in the homogeneous embedding space with the training corpus. Given a set of ophthalmic images, the extracted visual features are transformed to a compressed visual token and a sub-graph with relevant restored triples. 
Since the sub-graph is not guaranteed to be a completely accurate representation of the given images and natural noise exists in the clinical graph, we adopt a cross-modal encoder to encode the universal feature token and sub-graph information. To avoid influence from unrelated entities, a visible matrix is introduced during the cross-modal encoding process. Finally, reports are generated via a Transformer\cite{vaswani2017attention} decoder.

We conduct extensive experiments on the publicly available FFA-IR benchmark\cite{li2021ffa}. Experiments show that our CGT achieves the state-of-the-art performance of predicted reports under four automatic evaluation metrics and high AUC scores for the restored triples, providing a solid rationale for the explanation.

\begin{figure*}[t]
    \centering
    \includegraphics[width=0.95\textwidth]{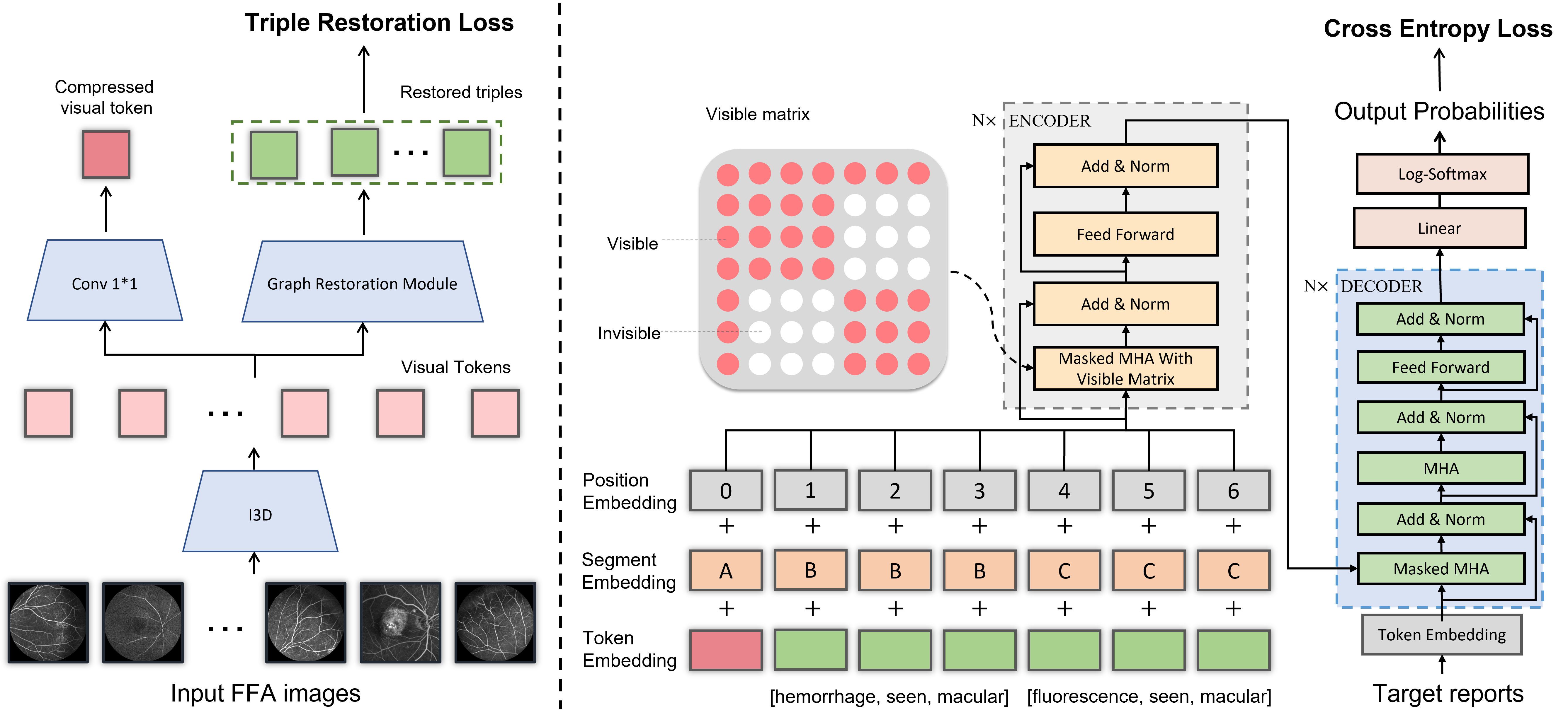}
    \caption{Illustration of our proposed cross-modal clinical graph transformer. Visual features extracted by an I3D are utilized to restore sub-graph information and compressed to one visual token. Then the cross-modal information encoded with visible matrix masked multi-head attention is used for report generation.}
    \vspace{-0.5cm}
    \label{fig:cgt}
\end{figure*}

\section{Related Work}

\subsection{Medical Report Generation}

Most existing medical report generation (MRG) models are proposed to describe radiology images, especially Chest X-Ray images~\cite{mimiccxr}, due to the limited access to medical resources. Recently, various medical report generation datasets have been released targeting on different medical modals, such as FFA images~\cite{li2021ffa}, lung CT scans~\cite{li2020auxiliary} and color fundus photography (CFP)~\cite{deepopht}, and attracted increasing attention. Inspired by work on image captioning, researchers have  adopted a hierarchical recurrent network (HRNN) to describe medical images at the beginning~\cite{jing2017automatic,xue2018multimodal}. In these HRNNs, the visual features extracted by a convolution neural network are attended with textual information to generate reports. Different to generic image captioning, there are data biases stemming from both visual and textual information. The textual data bias leads to similar sentences among different reports. Therefore, Cao \textit{et al.}~\cite{cao2018retrieve}, and Li \textit{et al.} summarized a set of sentence templates and used the retrieved semantic features to fill the templates and generate a report. With the success of Transformer in the vision-and-language field, many Transformer-based MRG models~\cite{emnlpMRG,li2020auxiliary,alfarghaly2021automated,nguyen2021automated} have been proposed to replace the LSTM since Transformer is one of the most effective encoder-and-decoder frameworks. Chen \textit{et al.} proposed a memory matrix to drive the decoding procedure. Alfarghaly \textit{et al.} introduced 105 tags and concatenated the weighted tag embedding with visual features for decoding. Unlike others, Wang \textit{et al.} firstly employs a selective search algorithm to extract the region-level image features to improve the MRG models. Since medical report generation is highly knowledge-driven, researchers have started incorporating medical knowledge to enhance the models.

\vspace{-0.2cm}

\subsection{Medical Knowledge Enhanced Models}

\vspace{-0.2cm}

In this section, we will introduce medical knowledge enhanced models for medical report generation and other medical domain tasks, medical QA, or memorization. The incorporated medical knowledge can be divided into three groups. 

The first kind is from radiologists' working patterns~\cite{li2020auxiliary, liu2021exploring}. In clinical practice, radiologists read images and write reports in a specific pattern to remind them of not missing any part of the images. After browsing the whole image, radiologists will focus on the suspicious regions. To make the model achieve this capability, Li \textit{et al.} introduced two kinds of auxiliary signals to guide the MRG model. Similarly, Liu \textit{et al.} adopted both posterior and prior knowledge to imitate the pattern with retrieved reports and a universal knowledge graph. Secondly, researchers explored the clinical knowledge behind the free-text reports to drive MRG models. Both \cite{li2019knowledge} and \cite{li2020auxiliary} extract normal and abnormal terminologies from corpus as nodes and automatically predict weights between these findings as edges to construct a knowledge graph. This graph worked as prior knowledge to drive the decoding procedure and restore a unique sub-graph for each case. In contrast, Zhang \textit{et al.}~\cite{zhang2020radiology} and Liu \textit{et al.} adopted an universal graph covering 20 findings to enhance the MRG models. In the last, the existing biomedical knowledge base is adopted to incorporate medical knowledge. The unified medical language system (UMLS)~\cite{bodenreider2004unified} is the largest biomedical knowledge base and is adopted in \cite{liu2020k} and \cite{he2019integrating} to enhance pretrained medical models for medical QA tasks. However, utilizing the existing knowledge base will bring in inconsistencies due to the heterogeneous embedding space arising from vocabulary and context mismatch. Since the entities and relations in UMLS are derived independently of the training corpus, when embedding node information, the embedded token vectors are inconsistent. Additionally, utilizing the full UMLS in MRG tasks will place a burden on the computation resources since it has $13,555,037$ triples, and most of them are irrelevant to our task.

\vspace{-0.35cm}

\section{Methodology}

\vspace{-0.15cm}

In this section, we introduce the clinical graph extraction scheme, and the process is shown in Figure~\ref{fig:scheme}. Then we detail the implementation of CGT (see Figure~\ref{fig:cgt} for the overall framework).

\vspace{-0.2cm}

\subsection{Notation}

\vspace{-0.15cm}
In ORG task, given a set of FFA images which represented by $I = \{x_1,x_2,\dots,x_{N_i}\}$, where $x_j$ and $N_i$ refer to the $j$-th FFA image and the number of total images, model is asked to generate a descriptive report encoded as $R = \{y_1,y_2,\dots,y_{Nr}\}$. While we denote the ground truth report by $\hat{R} = \{\hat{y}_1,\hat{y}_2,\dots,\hat{y}_{N\hat{r}}\}$. We extract entities and relations from all the training $\hat{R}$ to construct a clinical graph (CG), denoted as $\mathcal{G}$, which is a collection of triples $\epsilon = (e_s, r, e_o)$, where $e_s$ and $e_o$ denote the names of subjective and objective entities, and $r$ is the relation between them. All the triples are in CG, i.e., $\epsilon \in \mathcal{G}$. In this paper, English tokens are taken at the word-level and each token $y_i$, $e_i$ and $r_i$ are in the same vocabulary $\mathcal{V}$ whose size is $d_V$ to make all the embedding vectors consistent.
\vspace{-0.2cm}
\subsection{Clinical Graph Extraction Scheme}\label{sec:scheme}
\vspace{-0.15cm}
\begin{figure}
    \centering
    \includegraphics[width=0.45\textwidth]{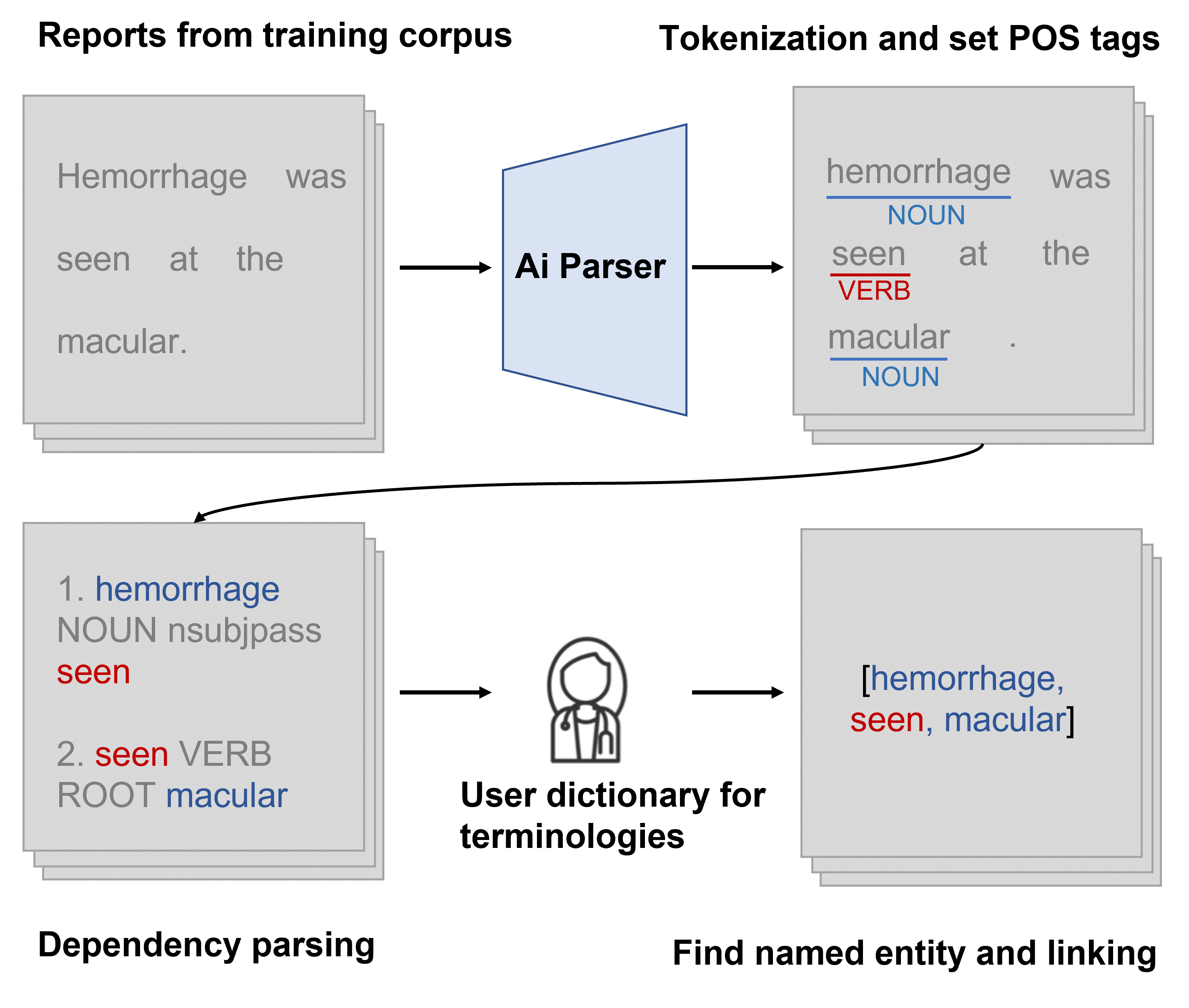}
    \caption{Process for extracting entities and relations from ophthalmic reports.}
    \vspace{-0.7cm}
    \label{fig:scheme}
\end{figure}

Recently, extracting clinical information from medical reports has received increasing attention\cite{jain2021radgraph, wu2021chest}.
The structured clinical information within the free-text reports is valuable for clinical reasoning and a variety of critical healthcare applications. We believe that ORG is one such application. However, due to the huge domain discrepancy between different medical models, transferring information from existing biomedical knowledge databases is unlikely to be effective.
In this subsection, we will introduce our information extraction scheme to detail how we construct a clinical graph $\mathcal{G}$ from ophthalmic reports. This scheme is implemented by a SpaCy\cite{spacy} natural language parser in an AI accelerating human-in-the-loop manner\cite{wu2020ai}. Notably, the ophthalmic reports used in this scheme are all derived from the training set to avoid target leakage.

To save the writing space, we take one sentence, ``\textit{Spotted obscured fluorescence (hemorrhage?) was seen at the inferior edge of the macular arch ring during left eye imaging.}'' from an ophthalmic report as an example, and the whole process is shown in Figure~\ref{fig:scheme}. Our scheme contains seven steps by following: \textbf{Tokenization}, taking the sentence into word-level and segmenting tokens into words, punctuation marks etc; \textbf{Part-of-speech tagging}, before automatically recognizing the relations between each pair tokens, we assign work types to each token, such as verb or nun; \textbf{Dependency parsing}, assigning syntactic dependency labels to describe the relations between individual tokens, such as `\textit{spotted}' is the attributive of subjective `\textit{fluorescence}'; \textbf{Lemmatization}, digging the base form of tokens. For example, the lemma of `was' is `is'; \textbf{Sentence boundary detection}, finding individual sentences to prevent the calculation across sentences; \textbf{Named entity recognition}, we create a user-dictionary to assist the machine in recognizing rare ophthalmic terminologies, such as `macular'; \textbf{Entity linking}, linking entities with their relation to creating triples. Triples extracted from the sample are ``\textit{fluorescence, seen, macular}'' and ``\textit{hemorrhage, seen, macular}'', respectively. Then we collect all the unique triples to construct the whole clinical graph $\mathcal{G}$. In total, our clinical graph contains $4,823$ triples, and more details are presented in Table~\ref{table:cg}.

\begin{table}
\centering
\caption{Statistics of our clinical graph.}\label{table:cg}
\begin{tabular}{ccc} 
\toprule
\# Entities & \# Relations & \# Triples  \\ 
\hline
        1,811    &       29       &        4,823     \\
\hline
\end{tabular}
\vspace{-0.5cm}
\end{table}

\subsection{Cross-modal Clinical Graph Transformer}\label{sec:cgt}

The traditional report generation models are based on an encoder-decoder architecture. Among all the encoder-decoder frameworks, Transformer\cite{vaswani2017attention} has achieved great success in various tasks. Therefore, we adopt a Transformer, the backbone of our proposed CGT, to describe ophthalmic images from the FFA-IR benchmark. As shown in Figure~\ref{fig:cgt}, our CGT is composed of a visual extractor, a graph construction module, a cross-modal encoder, and a language decoder.

\noindent\textbf{Visual Extractor} Different from describing radiology images, the average number of input images for each case is $97$ in the FFA-IR. Despite the benchmark proposed by \cite{li2021ffa} is adopting lesion features via a Faster-RCNN\cite{ren2015faster}, we utilize an I3D\footnote{\url{https://github.com/piergiaj/pytorch-i3d}} model pretrained on Kinetics\cite{i3d} to extract visual features from given images. Due to the reason that the entities in our CG contain both abnormalities and normal tissues, while the lesion information provided by the FFA-IR is all about the lesions or abnormalities. This data bias may mislead the message passing inter the CG.

Since the image numbers are different among each case, we first transform the given images and set a fixed length of $96$ for all the input images. For those whose length is more than $96$, we randomly down-sample some images. In contrast, we repeat the whole sequence until its length is $96$, when its length is below the threshold. The I3D model extracts one feature from every eight images, and the final visual features can be denoted as $f_V = \{f_1,f_2,\dots,f_{12}\}$, where $f_i \in \mathbb{R}^{12\times 1024}$. 

\noindent\textbf{Graph restoration module} The graph construction module is proposed to restore a sub-graph according to the visual features generated by the visual extractor. The sub-graph encoded as $\mathcal{G}_{s} = {\epsilon_1,\epsilon_2,\dots,\epsilon_{Ngs}}$ is a combination of triples. The whole process can be written as follows:
\begin{align}
    \mathcal{G}_{s} = max(0; BN(conv_{3\times 3}(f_V)))W_{f} + b_{f}
\end{align}
where $max(0;*)$ and BN represent the ReLU activation function and batch normalization operation, respectively; $W_f \in \mathbb{R}^{1024\times d_V}$ denotes learnable matrix
for linear transformation, while $b_{f}$ refers to the bias terms. Firstly, we adopt a convolution layer with a $3\times 3$ kernel followed by an operation sequence of batch normalization and ReLU activation to fuse the temporal information inside the $f_v$. Then the output has been projected by a linear transformation layer to the dimension of $d = d_V$. As mentioned, all the tokens in CG are in the same vocabulary with the training corpus; then, each vector is used to restore the index of entity or relation in $\mathcal{V}$. 

\noindent\textbf{Cross-modal Encoder} In this module, the visual features, and the graph information are encoded by self-attention mechanism\cite{vaswani2017attention}. The input of the cross-modal encoder comes from the visual extractor and the graph restoration module. As mentioned in \cite{liu2021exploring,wang2021self}, serve visual bias exists in most medical datasets for two reasons: the abnormal regions only take a small portion of the whole image, and the human tissues are highly similar. To alleviate the impact of visual bias, we compress the $f_v$ into one compressed visual token, encoded as $T_v \in \mathbb{R}^{d}$, and concatenate it with a sub-graph before fed to the embedding layer. The compressed visual token has two more advantages. Firstly, it promises that the sub-graph information is dominant to the input features. More importantly, it can be used to resist the inevitable noise inside the clinical graph adaptively since the knowledge graph can not be completely accurate. 

We utilize an `argmax' function on $\mathcal{G}_s$ and transform it into the one-hot format to represent the sub-graph, represented as $T_g = \{t_1,t_2,\dots,t_{Nt}|t_i \in \mathbb{R}^{d_v}\}$. 
After concatenation, we feed the cross-modal tokens, encoded as $T = \{T_v, T_g\}$, to the embedding layer. 
The function of the embedding layer is to convert the cross-modal tokens into embedding representations. Similar to the BERT\cite{devlin2018bert}, the embedding representation of CGT is the sum of three parts. 
Firstly, each token in $T_g$ is converted to an embedding vector of dimension $d = 512$ via a trainable lookup table. 
Different from BERT, the classification tag $[CLS]$ is replaced by $T_v$. Secondly, position embedding is added to the token embedding, and the formulation is written as follows:
\begin{align}
    PE_{pos,2i} & = \textit{sin}(\textit{pos}/1000^{2i/d}) \\
    PE_{pos,2i+1} & = \textit{cos}(\textit{pos}/1000^{2i/d})
\end{align}
where \textit{pos} is the position of each token, $i$ is the index of embedding dimension, and $d$ is the dimension of the hidden states. Lastly, segment embedding is employed to identify each sentence. Notably, we find that most sentences in the training corpus contain two triples. Therefore, we consider every six tokens as a sentence. The $T$ is marked with a sequence of segment tags, $\{A,B,\dots,B,C,\dots,C\}$, where $A$ represents the compressed visual token.

Then the embedded tokens are encoded by a cross-modal encoder, the whole process of an encoder layer can be written as:
\begin{align}
    f_e(t) & = BN(FFN(e_{attn})+e_{attn}) \\
    e_{attn} & = BN(MMHA(t)+t)
\end{align}
Where $\textit{FFN}$ represents the feed forward layer, and \textit{MMHA} represents the mask multi-head attention. The feed forward layer contains two linear layers with ReLU activation. It makes sure the dimensions of transformer input and output are the same. Another difference between our CGT and Transformer is that we adopt \textit{MMHA} instead of \textit{MHA} during the encoding process and introduce a visible matrix, $M_v$, to limit the impact of unrelated triples. The computation between unrelated triples is useless and untrue, which may also lead the changes to the original meanings. The visible matrix is presented in Figure~\ref{fig:cgt}, and it can limit the message passing inter the sentence or between the universal token. The \textit{MMHA} can be written as:
\begin{align}
    \mathbf{h}_i^t = \text{softmax}(\frac{\mathbf{Q}_i(\mathbf{K}^t)^{'} M_v}{\sqrt{d}})\mathbf{V}^t
\end{align}
where $\{ \mathbf{Q}, \mathbf{K^*}, \mathbf{V^*}\}$ are the packed $d$-dimensional \textit{Query, Key, Value} vectors. 

\noindent\textbf{Language Decoder} We adopt the vanilla Transformer decoder as our language decoder. The whole process of a decoder layer can be written as:
\begin{align}
    f_d(\mathbf{y}) & = BN(FFN(e_{c_attn})+e_{c_attn}) \\
    e_{c_attn} & = BN(MHA(e_{attn}, f_e(\mathbf{t}))+e_{attn}) \\
    e_{attn} & = BN(MMHA(\mathbf{y})+\mathbf{y}))
\end{align}
where \textit{MMHA} represents the original masked multi-head self-attention, $\mathbf{y}$ is the input of decoder and $y_t$ is the $t-$th input token in time step $t$. Cross-attention sublayer receives the output of encoder $f_e(\mathbf{t})$ and previous sublayer $e_{attn}$.
In where, for each head, $\{ \mathbf{Q}, \mathbf{K^*}, \mathbf{V^*}\}$ comes from $\mathbf{Q} = W_{q}*e_{attn}$, $\mathbf{K} = W_{k}*f_e(\mathbf{x})$, and $\mathbf{V} = W_{v}*f_e(\mathbf{x})$, where $W_{*}$ is the weight of a Linear layer. The $f_d(\mathbf{y})$ will be sent to a Linear $\&$ Log-Softmax layer to get the output of target sentences. Notably, only token embedding is adopted during the decoding procedure. The entire recursive generation process can be written as follows:
\begin{align}
    p(\hat{R}|I) = \prod_{t=1}p(\hat{y}_t|\hat{y}_1,\dots,\hat{y}_{t-1}, I)
\end{align}

\noindent\textbf{Objectives} Similar to the image captioning tasks, existing medical report generation approaches adopt cross-entropy loss to evaluate the differences between the predicted and the ground truth reports at the word level. Although many works attempt to explore auxiliary signals to drive the report generation, these signals can not supervise the learning process. For example, Li \textit{et al.}\cite{li2020auxiliary} introduced an internal visual signal to locate the abnormal regions. However, there is no ground truth for the abnormal region bounding. Similarly, the accurate weights correlated paired findings in \cite{zhang2020radiology,li2019knowledge} are also unavailable. Therefore, the effect of auxiliary signals has been limited. 

In this paper, we additionally introduce a triple restoration loss~\cite{he2019integrating} to supervise the sub-graph restoration process since our clinical graph extraction scheme provides the ground truth structured information. It guarantees that the accurate graph information will be encoded with the visual features for report generation and is also what makes this method so effective. Specifically, the total loss function used in our CGT can be written as follows:
\begin{align}
    \mathcal{L}_{RG} = \lambda_{CE}\mathcal{L}_{CE} + \lambda_{TR}\mathcal{L}_{TR}
\end{align}
where $\lambda_{CE}$ and $\lambda_{TR}$ are hyper-parameters balancing two terms. The first loss term $\mathcal{L}_{CE}$ is the cross-entropy loss. The second loss term is the triples restoration loss function to measure the energy of a knowledge triple. The specific process can be written as follows:
\begin{align}
    \mathcal{L}_{TR} = \sum\limits_{\epsilon \in \mathcal{G}}\max(d(\epsilon)-d(f(\epsilon))+\gamma, 0)
\end{align}
where $\epsilon = (e_s,r,e_o)$, $d(\epsilon) = |e_s+r-e_o|$, $\gamma>0$ is a margin hyper-parameter, $f(\epsilon)$ is an entity replacement operation that the subjective or objective entity in a triple is replaced and the replaced triple is an invalid triple in $\mathcal{G}$. Here, $e_s$,$e$ and $e_o$ are the indexes of the subjective, relation and objective tokens in $\mathcal{V}$.

\section{Experiments}\label{sec:experiments}

\subsection{FFA-IR Benchmark}\label{sec:dataset}

In this paper, we adopt a recently released largest ophthalmic report generation dataset to date, i.e., FFA-IR~\cite{li2021ffa}, to verify the effectiveness of our approach. FFA-IR contains $10,790$ reports and $1,048,584$ FFA images. In addition, FFA-IR provides bilingual reports for each case and $12,166$ lesion bounding information, which can explain the diagnosis process. For a fair comparison, we report our results on the official splits, in which 8,016 reports/99,161 images for training, 1,069 reports/93,274 images for validation, and 1,604 reports/138,026 images for testing. Similar to the settings in Section~\ref{sec:scheme}, all the tokens are converted into lower cases, and those whose frequency of occurrence is less than three are removed, resulting in $3,241$ tokens including both words and marks. We additionally add $[PAD]$, $[SOS]$, $[EOS]$ and $[UNK]$ tags whose indexes are 0, 1, 2 and 3 into the vocabulary, resulting in $3,245$ tokens in $\mathcal{V}$. In addition, FFA-IR is the only publicly available ORG dataset based on our knowledge.

\subsection{Baseline, Evaluation Metrics, and Settings}

Along with the dataset, three benchmark models, which adopt spatial features via a ResNet\cite{resnet}, temporal features via an I3D, and lesions features via a Faster-RCNN, have been released. In this paper, we adopt the combination of I3D and Transformer as our baseline method for two reasons. On the one hand, the lesion information does not explain normal tissues, which take a considerable portion in our CG's nodes; On the other hand, the critical information inside the spatial features are easily inundated by global features during the feature fusion process~\cite{li2021ffa}.

We adopt two kinds of metrics to validate the effectiveness of this method. The widely used natural language generation (NLG) metrics\footnote{\url{https://github.com/tylin/coco-caption}}, including BLEU~\cite{papineni-etal-2002-bleu}, CIDEr~\cite{vedantam2015cider}, METEOR~\cite{banerjee2005meteor} and ROUGE-L~\cite{lin-2004-rouge}, are adopted to evaluate the quality of predicted reports in word-level. Cider is adopted as the main metric since BLEU and METEOR are mainly used for machine translation evaluation, and ROUGE-L is designed for summaries. We also figure the micro-average of receiver operation characteristic (ROC) curve and report the area under the ROC curve (AUC) to evaluate the accuracy of the restored sub-graph.

The whole network is implemented by Pytorch~\cite{paszke2019pytorch} based on Python 3.7 and trained on two GeForce RTX 2080Ti GPUs. The images are resized to $224\time 224$ before being fed into the I3D, and the batch size is $8$. The maximum length of $T$ is 90, padded with tag $[PAD]$. The embedding space for both visual and graph tokens is $512$, and the dimension of the hidden states in the Transformer is also $512$. Both encoder and decoder consist of six blocks and 8 heads. The ADAM~\cite{kingma2014adam} is utilized for optimizing all the parameters in our CGT, and the learning rate is $1e-4$. The whole network is trained for 50 epochs. We adopt greedy decoding when testing. 

\subsection{Main Results}\label{sec:main_res}

\begin{table*}[t]
\centering
\caption{The results of NLG metrics of our proposed CGT and other state-of-the-art methods on the FFA-IR dataset. Bold numbers denote the best performance in their columns.}\label{table:main_results}
\begin{tabular}{lcccccccc} 
\toprule
Methods   & Year & BLEU-1         & BLEU-2         & BLEU-3         & BLEU-4         & METEOR         & ROUGE-L        & CIDEr           \\ 
\hline
CoAtt\cite{jing2017automatic}     & 2018 & 0.313          & 0.200          & 0.144          & 0.111          & 0.197          & 0.247          & 0.254           \\
Show-Tell\cite{vinyals2015show} & 2015 & 0.306          & 0.197          & 0.142          & 0.109          & 0.191          & 0.247          & 0.232           \\
Top-Down\cite{anderson2018bottom}  & 2018 & 0.320          & 0.217          & 0.162          & 0.127          & 0.207          & 0.289          & 0.363           \\
Gounded\cite{gounded}   & 2020 & 0.396          & 0.319          & 0.261          & 0.218          & \textbf{0.229} & \textbf{0.353} & 0.576           \\
AdaAtt\cite{adaatt}    & 2017 & 0.292          & 0.181          & 0.127          & 0.095          & 0.205          & 0.236          & 0.234           \\
R2Gen\cite{emnlpMRG}     & 2020 & 0.330          & 0.225          & 0.167          & 0.132          & 0.210          & 0.296          & 0.367           \\ 
\hline
I3D+T\cite{li2021ffa}     & 2021 & 0.428          & 0.341          & 0.276          & 0.229          & 0.213          & 0.334          & 0.561           \\
Faster+T\cite{li2021ffa}  & 2021 & 0.443          & 0.355          & 0.288          & 0.240          & 0.205          & 0.341          & 0.590           \\ 
\hline
CGT       & Ours & \textbf{0.456} & \textbf{0.363} & \textbf{0.295} & \textbf{0.243} & 0.227          & 0.345          & \textbf{0.599}  \\
\bottomrule
\end{tabular}
\end{table*}

\begin{figure}
    \centering
    \includegraphics[width=0.45\textwidth]{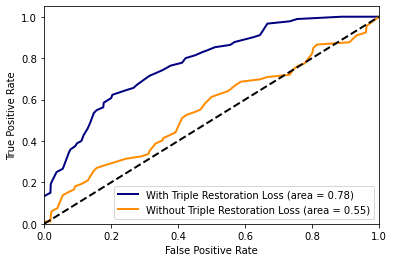}
    \caption{Micro-average of receiver operating characteristic curve for sub-graph restoration.}
    \label{fig:roc}
    \vspace{-0.5cm}
\end{figure}

\noindent\textbf{Report generation} In Table~\ref{table:main_results}, we compare our CGT with a wide range of existing models. I3D+T~\cite{li2021ffa} and Faster+T~\cite{li2021ffa} are the two benchmark models achieving the state-of-the-art performance on FFA-IR dataset. R2Gen~\cite{emnlpMRG} and CoAtt~\cite{jing2017automatic} are the state-of-the-art radiology report generation models. The remaining presented works are from image captioning approaches. As shown in Table~\ref{table:main_results}, our CGT outperforms the state-of-the-art method across all NLG metrics. The improved performance of CGT demonstrates the validity of our practice in incorporating prior medical into ophthalmic report generation.

\noindent\textbf{Sub-graph restoration} In Figure~\ref{fig:roc}, we show the micro-average of ROC for sub-graph restoration and present the AUC scores when the proposed model is trained with triple loss restoration loss or not. With the triple restoration loss, the AUC score increased from 0.55 to 0.78 significantly. This improvement demonstrates the effectiveness of triple restoration loss and the accuracy of our restored sub-graph. Without the triple restoration loss, the restored sub-graph is similar to a sequence of random triples. It also verifies the importance of our clinical graph extraction scheme.

\subsection{Quantitative Analysis}

In Table~\ref{table:q_a}, we present the results of quantitative analysis to investigate the contribution of each component in our CGT. The baseline model is a combination of I3D and Transformer proposed by \cite{li2021ffa}.

\begin{table*}
\centering
\caption{Quantitative analysis and human study of proposed method, where CVT, VM and TRL are the short for compressed visual token, visible matrix and triple restoration loss, respectively.}\label{table:q_a}
\begin{tabular}{c|ccccc|cccc|c} 
\toprule
Settings & I3D & Triples    & CVT & VM & TRL & CIDEr          & BLEU-4         & ROUGE          & METEOR & Hit($\%$)   \\ 
\hline
Baseline     & \checkmark   &        &     &    &     & 0.561          & 0.229          & 0.334          & 0.213     & 21.6      \\    
\hline
(a)      &     & \checkmark      &     &    &     & 0.223          & 0.087          & 0.218          & 0.200    & -       \\
(b)      &     & Random &     &    &     & 0.223          & 0.085          & 0.220          & 0.204       & -    \\
(c)      &     & \checkmark      &     &    & \checkmark   & 0.561          & 0.226          & 0.287          & 0.209     & -      \\
(d)      &     & \checkmark      &     & \checkmark  & \checkmark   & 0.569          & 0.231          & 0.309          & \textbf{0.228}  & -\\
(e)      &     & \checkmark      & \checkmark   &    & \checkmark   & 0.586          & 0.240          & 0.332          & 0.225       & -     \\
(f)      & \checkmark   & \checkmark      &     &    & \checkmark   & 0.573          & 0.242          & 0.324          & 0.226       & -     \\ 
\hline
CGT      &     & \checkmark      & \checkmark   & \checkmark  & \checkmark  & \textbf{0.599} & \textbf{0.243} & \textbf{0.345} & 0.227      & \textbf{44.7}     \\
\bottomrule
\end{tabular}
\vspace{-0.5cm}
\end{table*}

\begin{figure*}
\centering
\includegraphics[width=\textwidth]{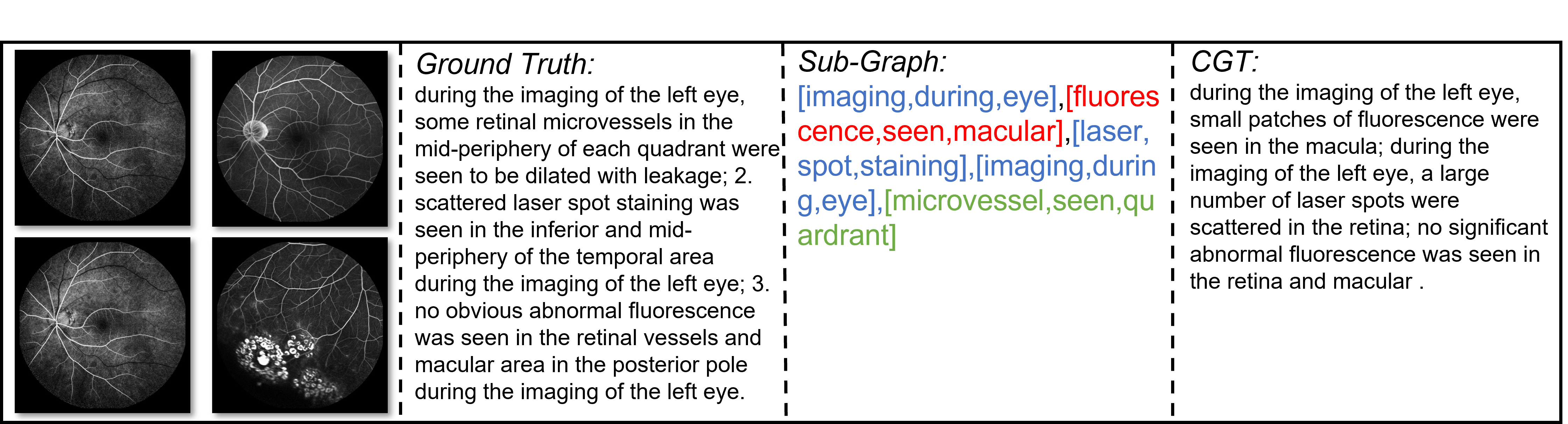}
\caption{Illustrations of reports from the ground truth and CGT, and the restored sub-graph. The blue, red, and greed triples represent the true positive, false positive, and false negative.}
\vspace{-0.8cm}
\label{figure:vs}
\end{figure*}

\noindent\textbf{Effect of clinical graph} In this section, we evaluate the effectiveness of the proposed clinical graph, including triples and triples restoration loss.

Comparing the results in baseline and (a) in Table~\ref{table:q_a}, we can find that without the triple restoration loss, the automatically restored sub-graph fails to drive the model to generate an accurate report. In (b), we randomly restore a sub-graph instead of based on the input visual features. Along with the AUC scores in Figure~\ref{fig:roc}, these demonstrate that only the relevant and accurate prior knowledge can improve the effectiveness of diagnostic models. Encouragingly, Table~\ref{table:q_a} Baseline and (c) illustrates that the results of utilizing the clinical graph only are competitive to the baseline. These results verify that the triples restoration loss can supervise the sub-graph restoration process and guarantee the accuracy of the incorporated prior knowledge.

\noindent\textbf{Effect of visible matrix} Visible matrix is another essential component in our CGT. This concept is widely used in knowledge-enhanced pretraining works~\cite{he2019integrating,liu2020k} with various formulations. In this paper, the visible matrix is adopted during the cross-modal encoding process for two purposes. On the one hand, we hope it can limit the impact of unrelated triples; On the other hand, we want the message can pass between the visual features and each triple.

The results between (c) and (d), (e) and CGT in Table~\ref{table:q_a} demonstrate the effectiveness of the visible matrix. We can see that the performances increase substantially when integrating visible matrix with (c) and (e), e.g., 0.561 $\rightarrow$ 0.569 and 0.586 $\rightarrow$ 0.599 in CIDEr score. Firstly, by comparing the results of (c) and (d), the visible matrix limits the impact from unrelated triples and greatly enhances the information interaction between related triples. Therefore, we speculate that the entity and relation representations can be well trained and improve the quality of predicted reports. When working in CGT, the visible matrix additionally facilitates the message passing between the visual features and each triple. There is inevitable noise among the knowledge graph since the relation is not a `hard' label. Although triple representations can be well learned, the triple may not be relevant to the input case. Therefore, the visual features play a role in de-noise adaptively. Furthermore, the visible matrix makes sure that the cross-modal signals can interact with each other. 

\vspace{-0.1cm}
\noindent\textbf{Effect of compressed visual token} The effectiveness of the compressed visual token is verified when comparing the results of (c), (e), and (f) in Table~\ref{table:q_a}. As discussed, there are always noises existing in a knowledge graph. Therefore, one of the purposes for proposing a compressed visual token is to keep the accurate signals from original meanings when the sub-graph is inaccurate. When integrating the compressed visual token, the quality of predicted reports improves significantly comparing (c) and (e) and outperforming the baseline method. It demonstrates the importance of visual signals in the $T$. We also conducted an experiment to compare the performances of injecting prior knowledge into the compressed visual token and temporal features ((e) and (f)). We can find that the performances decrease slightly when using all the temporal features, e.g., 0.586 $\rightarrow$ 0.573 in the CIDEr score. We speculate the reason is that too many visual tokens will impair the effectiveness of prior knowledge. Therefore, using the compressed visual token can make the prior knowledge dominant. Notably, the visible matrix is modified when using all temporal features. 

\noindent\textbf{Human study} In this section, we invited three senior ophthalmologists to evaluate the quality of predicted reports by the baseline model and our CGT. As shown in Table~\ref{table:q_a}, ophthalmologists regarded that $44.7\%$ of predicted reports by CGT can describe the given FFA images more accurately. The human study results demonstrate that our CGT outperforms the baseline model in both NLG metrics and clinical practice. Ophthalmologists also mentioned that there were $33.7\%$ of predicted reports by both methods that failed to describe any key finding.
\vspace{-0.25cm}
\subsection{Qualitative Analysis}
\vspace{-0.25cm}
In this section, we conduct qualitative analysis for better understanding our approach via an intuitive example. Given a set of input FFA images, our CGT first restores a sub-graph which is further incorporated with visual features to generate a report.

As shown in Figure~\ref{figure:vs}, one restored sub-graph consists of four triples, and each triple describes a relation between the subjective and objective entity, e.g., \textit{[fluorescence,seen,macular]} represents that based on the prior clinical knowledge, \textit{'fluorescence'} can be seen in the \textit{'macular}. The number of triples is depended on the length of the input FFA images. Among the restored triples, \textit{[fluorescence,seen,macular]} is the false positive triple which leads to the incorrect sentence \textit{during the imaging of the left eye, small patches of fluorescence were seen in the macular.} This phenomenon shows that our CGT is capable of extending triples to a relevant sentence. Notably, due to the serve textual bias among the training corpus, the sub-graph restoration also suffers since the clinical graph is constructed from the training corpus. \textit{[fluorescence,seen,macular]} is one of the bias triples and exists in $92\%$ training samples. Accurately restored the triple \textit{[laser,spot,staining]} verifies the effectiveness of our CGT to detect abnormalities among the input images and translate them into sentences. It also demonstrates that our CGT is highly capable in sub-graph restoration owing to the triple restoration loss. The last predicted sentence is not relevant to any triple in the restored sub-graph. However, this information can be provided by the compressed visual token.

\vspace{-0.35cm}
\section{Conclusion and Discussion}
\vspace{-0.2cm}
In this paper, we present an effective cross-modal clinical graph transformer for ophthalmic report generation. To obtain prior medical knowledge, we propose an information extraction scheme to construct a clinical graph from ophthalmic reports. The prior knowledge inside this graph is further restored to a sub-graph which is injected into the visual features for report generation. The experiments and analyses on the public FFA-IR dataset support our arguments and verify the effectiveness of our approach. Along with achieving state-of-the-art performances, the restored sub-graph also improves the explainability of our approach.

\noindent\textbf{Negative societal impact} As with other automatic diagnostic methods, our algorithm should be utilized carefully in clinical practice since medical decisions may lead to significant consequences, including death. Therefore, while our AI diagnostic method can provide a strong rationale for judgment along with satisfactory performances, it should only be used as an auxiliary resource.

\noindent\textbf{Limitations} Our clinical graph is constructed in an automatic manner from a training corpus; therefore, we cannot guarantee the complete accuracy of our graph. We are inviting more experienced ophthalmologists to verify this graph. In addition, our method is not sufficiently general to support other report generation tasks. For each task, we will need to update the information extraction methods and construct a new clinical graph.

\vspace{-0.4cm}
\section*{Acknowledge}
\vspace{-0.25cm}
This work is partially supported by Australian Research Council (ARC) Discovery Early Career Researcher Award (DECRA) under DE190100626, Guangdong Province Basic and Applied Basic Research (Regional Joint Fund-Key) Grant No.2019B1515120039, Guangdong Outstanding Youth Fund (Grant No. 2021B1515020061). We would like to acknowledge Prof.Feng Wen from Fundus Department, Zhongshan Ophthalmic Center, Sun Yat-Sen University for sharing data.

%%%%%%%%% REFERENCES
{\small
\bibliographystyle{ieee_fullname}
\bibliography{main}
}

\end{document}